\documentclass[conference]{IEEEtran}

\usepackage[table,xcdraw]{xcolor}

\usepackage{url}
\usepackage[hidelinks]{hyperref}
\usepackage{enumitem}
\usepackage{graphicx}   % Only load once (removed duplicate)
\usepackage{cite}
\usepackage{float}
\usepackage{csquotes}
\usepackage{adjustbox}
\usepackage{fancyvrb}
\usepackage{multirow}
\usepackage[most]{tcolorbox} % Load AFTER xcolor

\tcbset{
  answerbox/.style={
    colback=gray!10,
    boxrule=1.5pt,
    arc=10pt,
    left=7pt,
    right=7pt,
    top=7pt,
    bottom=7pt,
    rounded corners
  }
}

\begin{document}

\title{Automatic Classification of User Requirements from Online Feedback - A Replication Study}

% \author{
% \IEEEauthorblockN{Meet Bhatt}
% \IEEEauthorblockA{\textit{Schulich Sch. of Eng.} \\
% \textit{University of Calgary}\\
% Calgary, AB, Canada \\
% meet.bhatt@ucalgary.ca}
% \and
% \IEEEauthorblockN{Nic Boilard}
% \IEEEauthorblockA{\textit{Schulich Sch. of Eng.} \\
% \textit{University of Calgary}\\
% Calgary, AB, Canada \\
% nicole.boilard@ucalgary.ca}
% \and
% \IEEEauthorblockN{Muhammad Rehan Chaudhary}
% \IEEEauthorblockA{\textit{Schulich Sch. of Eng.} \\
% \textit{University of Calgary}\\
% Calgary, AB, Canada \\
% muhammadrehan.chaudh@ucalgary.ca}
% \and
% \IEEEauthorblockN{Cole Thompson}
% \IEEEauthorblockA{\textit{Schulich Sch. of Eng.} \\
% \textit{University of Calgary}\\
% Calgary, AB, Canada \\
% cole.thompson2@ucalgary.ca}
% }

\author{Meet Bhatt, Nic Boilard, Muhammad Rehan Chaudhary, Cole Thompson, \\ Jacob Idoko, Aakash Sorathiya, Gouri Ginde \\

% \IEEEauthorblockA{{
Dept. of Electrical and Software Engineering, University of Calgary, Canada \\
\{meet.bhatt, nicole.boilard, muhammadrehan.chaudh, cole.thompson2, \\ jacob.idoko, aakash.sorathiya, gouri.ginde\}@ucalgary.ca}

\maketitle

\begin{abstract}
Natural language processing (NLP) techniques have been widely applied in the requirements engineering (RE) field to support tasks such as classification and ambiguity detection. Although RE research is rooted in empirical investigation, it has paid limited attention to the replication of NLP for RE (NLP4RE) studies. Additionally, the rapidly advancing realm of NLP is creating new opportunities for efficient, machine-assisted workflow applications, which can bring new perspectives and results to the forefront. Thus, in this study, we replicate and extend a previous NLP4RE study (baseline), ``Classifying User Requirements from Online Feedback in Small Dataset Environments using Deep Learning", which evaluated different deep learning models for requirement classification from user reviews. In this study, we reproduced the original results using the publicly released source code, thereby helping to strengthen the external validity of the baseline study. We then extended the baseline setup by evaluating the model's performance on an external (new) dataset and comparing the results to a GPT-4o zero-shot classifier. Furthermore, we prepared the replication study ID-card for the baseline study, which is an important aspect to evaluate replication readiness. 

The results showed diverse reproducibility levels across different models, with Naive Bayes demonstrating perfect reproducibility. In contrast, BERT and other models showed mixed results. Our findings also revealed that baseline deep learning models, BERT and ELMo, exhibited good generalization capabilities on an external dataset, and the GPT-4o model showed performance comparable to traditional baseline machine learning models. Additionally, our assessment of the replication study ID-card confirmed the replication readiness of the baseline study; however, the missing environment setup files would have further enhanced the readiness. We include this missing information in our replication package and provide the replication study ID-card for our study to further encourage and support the replication of our study.
\end{abstract}


\section{Introduction}
\textit{Replication} is an important aspect of empirical evaluation that involves repeating an experiment under similar conditions using a different subject population \cite{abualhaija2024replication}. Replicability is currently regarded as a major quality attribute in software engineering (SE) research, and it is one of the main pillars of Open Science \cite{mendez2020open}. It allows us to build knowledge about which results or observations hold under which conditions and confirm or refute hypotheses and previous results \cite{khatwani2017advancing}.

The key distinction in SE replication studies is between \textit{internal} and \textit{external} replications \cite{brooks2008replication} \cite{ginde2025so} \cite{ruwanpura2023automatic}. Internal replication is conducted by the original researchers, while external replication is carried out by independent researchers. Brooks et al. \cite{brooks2008replication} emphasized the importance of external replication for validating SE principles and guidelines. However, external replication is still rare in RE. Although the number of software engineering replications was updated from 20 in Sj{\o}berg et al.’s survey \cite{sjoberg2005survey} to 133 in da Silva et al.’s study \cite{da2014replication}, 31 of the 32 RE replications (97\%) were internal ones. Furthermore, replication does not appear to be commonly practiced in the natural language processing for requirements engineering (NLP4RE) research strand, despite its growing interest \cite{abualhaija2024replication} \cite{ginde2025so} \cite{ruwanpura2023automatic}.

% Probably the most well-known distinction in software engineering is between \textit{internal} and \textit{external} replications, as defined in \cite{brooks2008replication}. Internal replication is undertaken by the original researchers themselves or the team involving them, whereas external replication is performed by independent researchers. Brooks et al. \cite{brooks2008replication} pointed out that, without the confirming power of external replication, many principles and guidelines in software engineering should be treated with caution. Unfortunately, external replication is still rare in RE. Although the number of software engineering replications was updated from 20 in Sj{\o}berg et al.’s survey \cite{sjoberg2005survey} to 133 in da Silva et al.’s study \cite{da2014replication}, 31 of the 32 RE replications (97\%) were internal ones. Furthermore, replication does not appear to be commonly practiced in the natural language processing for requirements engineering (NLP4RE) research strand, although it is receiving a lot of attention from both industry practitioners and academic researchers \cite{abualhaija2024replication}.

Therefore, to address this research gap, we conducted the replication of a previous study (baseline) by Mekala et al. \cite{Mekala2021} in the NLP4RE domain, as part of our undergraduate research project for Software Requirements Engineering Winter term course at the Dept. of Electrical Engineering, University of Calgary. Replicating a study in this course is also part of another pedagogical study \cite{gindeEASE}, which explores teaching professional ethics using a replication study as a tool - the ethics for this study were reviewed and approved by our university's
Institutional Research Information Services Solution (IRISS) Board, reference \#REB23-1414, University of Calgary. 

We extended this replication study further and conducted additional experiments to compare the proposed NLP technique with the GPT-based zero-shot classifier and utilized a different dataset to verify the generalizability of the proposed NLP technique. None of the original authors were part of this replication study; we contacted the lead author once to receive information about the baseline study in the initial stages and received the revised dataset; however, we did not use this revised dataset in our study as it was not directly comparable due to the difference in dataset size.

\textbf{Baseline study:} In this study, we replicate Mekala et al.'s study \cite{Mekala2021} titled ``Classifying User Requirements from Online Feedback in Small Dataset Environments using Deep Learning", published in at IEEE 29th International Requirements Engineering conference in 2021. The primary objective of the baseline study was to evaluate the performance of different deep learning (DL) algorithms for classifying requirements from user reviews (binary classification). For this purpose, they leveraged the labeled dataset provided by Van Vliet et al. \cite{van2020identifying} and fine-tuned three DL models, including FastText, Embeddings from Language Models (ELMo), and Bidirectional Encoder Representations from Transformers (BERT), on this dataset. They also considered two traditional machine learning (ML) models, Term Frequency-Inverse Document Frequency (TF-IDF) with Support Vector Machine (SVM) and Naive Bayes, as performance benchmarks for DL models.

\begin{table*}[!htpb]
\centering
\renewcommand{\arraystretch}{1.2}
\caption{Statistical information for the 3 datasets used in this study}
\label{tab:dataset_overview}
\begin{tabular}{lrcccp{5.5cm}}
\textbf{Dataset} & \textbf{\# Samples} & \textbf{Helpful (1) / Useless (0)} & \textbf{Avg Length} & \textbf{Std Dev} & \textbf{Example Review} \\
\hline
P1 Baseline         & 1000  & 48.4\% / 51.6\% & 19.9 & 22.3 & ``Crashes during video calls need urgent fix'' \\
P2 Baseline         & 1242  & 45.3\% / 54.7\% & 12.9 & 9.8  & ``Would pay for dark mode option'' \\
% P1 Gold Standard   & 1000  & 55\% / 45\% & 19.9 & 22.3 & ``Update broke my saved preferences'' \\
% P2 Gold Standard   & 1340  & 61\% / 39\% & 11.2 & 8.1  & ``Why no tablet version yet?'' \\
External/Additional Dataset   & 5068  & 37.6\% / 62.4\% & 18.2 & 19.9 & ``Battery drain improved in latest update'' \\
\end{tabular}
\end{table*}

We formulate and evaluate the following four research questions (RQs) in this study:

\textbf{RQ1: (Sanity check) To what extent was the baseline study (Mekala et al. \cite{Mekala2021}) replicable?}
The replication of outcomes from an open-source program and tool to reproduce the findings presents considerable challenges owing to code dependencies and system configurations. Addressing these obstacles, the reproduction and assessment of the baseline study outcomes could significantly enhance the external validity of the findings. This highlights the importance of transparency in research, encompassing the availability of datasets, guidelines for annotation, preprocessing methodologies, model configurations, and evaluation metrics.

\textbf{RQ2: (Generalizability) How does the baseline study design perform for an external dataset (from Zaeem et al. \cite{Zaeem2023mohammadzaeem})?}
Evaluating the baseline study design on an external dataset helps determine whether the proposed NLP technique can generalize and maintain its performance outside the initial test conditions. This enables the replication study to validate the practical applicability of the original findings and identify limitations or required modifications to enhance the proposed method. To this aim, we utilize a similar dataset provided by Zaeem et al. \cite{Zaeem2023mohammadzaeem}. 

\textbf{RQ3: (Extension) Can a GPT-based zero-shot classifier match or outperform fine-tuned models proposed by the baseline study?}
The baseline study has proposed state-of-the-art supervised classification models that are fine-tuned on the labeled dataset. However, curating a ground truth dataset requires manual efforts and is time-consuming. Therefore, we investigate to what extent we can leverage GPT-based zero-shot classifiers, which do not require any fine-tuning, to classify user requirements from online user feedback automatically. We compare the performance of the GPT-based zero-shot classifier with the fine-tuned classifiers provided in the baseline study.

\textbf{RQ4: (Replication readiness) To what extent are the baseline and our study replication ready?}
While there is a lot of emphasis on the need for more replication studies to foster the reproducibility and external validity of exponentially evolving NLP methodologies in RE, there is also a need to evaluate the studies for their replicability readiness. Thus, we evaluate the baseline study for such replication readiness using the ID-card provided by Abualhaija et al. \cite{abualhaija2024replication} and try our best to address the shortcomings in this replication study package further.

We make the following contributions through this study:

\begin{itemize}
    \item Utilizing the source code and the dataset repository from the baseline study, Mekala et al. \cite{Mekala2021}, a RE conference main track paper, published in 2021, we reproduced and validated the results originally published in this work.
    \item We further evaluated their methodology on a separate dataset provided by Zaeem et al. \cite{Zaeem2023mohammadzaeem} to explore the generalizability of the baseline study. As such, we executed their code and evaluated methodology on an external dataset \cite{Zaeem2023mohammadzaeem}, thus exploring the external validity of the baseline study.
    \item We extended the baseline study by integrating the GPT-based zero-shot classifier to support requirement classification in scenarios with limited or no labeled data.
    \item We evaluated the baseline study for its replication readiness using the ID-card defined by Abulhaija et al. \cite{abualhaija2024replication}. Also, we tried (our best) to capture some of these missing elements in our replication package.
    \item Similar to the baseline study, we make our source code and replication package publicly available\footnote{\url{https://doi.org/10.5281/zenodo.15612003}}. 
\end{itemize}

The rest of the paper is structured as follows. Section \ref{data} describes the datasets, and Section \ref{design} presents the study design. Results are explained in Section \ref{results}, followed by threats to validity in Section \ref{threats}. Finally, Section \ref{rw} reviews the related work and Section \ref{conclude} provides the concluding remarks and future directions.

\section{Dataset} \label{data}

In this section, we discuss two datasets used for evaluating our RQs. Dataset from the baseline study \cite{Mekala2021} available on figshare\footnote{\url{https://figshare.com/articles/dataset/data_and_code_zip/14273594}} is referred to as the baseline benchmark dataset. The dataset used for evaluating the generalizability of the baseline study is from Zaeem et al. \cite{Zaeem2023mohammadzaeem}, referred to as an additional (external) dataset.
Table~\ref{tab:dataset_overview} shows statistical information from all three datasets used in our study: the P1 and P2 datasets from the baseline study \cite{Mekala2021}, and the external dataset used for testing model generalizability \cite{Zaeem2023mohammadzaeem}. These include: the number of samples in each dataset, the distribution of helpful versus useless reviews/sentences, the average length (mean number of words per review or sentence), the standard deviation of text lengths, and example reviews from each dataset.
% This study uses two main datasets: the baseline benchmark dataset from Mekala et al.\cite{Mekala2021} available on figshare\footnote{\url{https://figshare.com/articles/dataset/data_and_code_zip/14273594}}, and the extended dataset to address RQ2 and RQ3. The baseline dataset supports replication (RQ1), while the extended dataset is used to evaluate model generalizability and zero-shot performance. 
\subsection{Baseline Dataset}
\begin{figure}[ht]
    \centering
    \includegraphics[width=\linewidth]{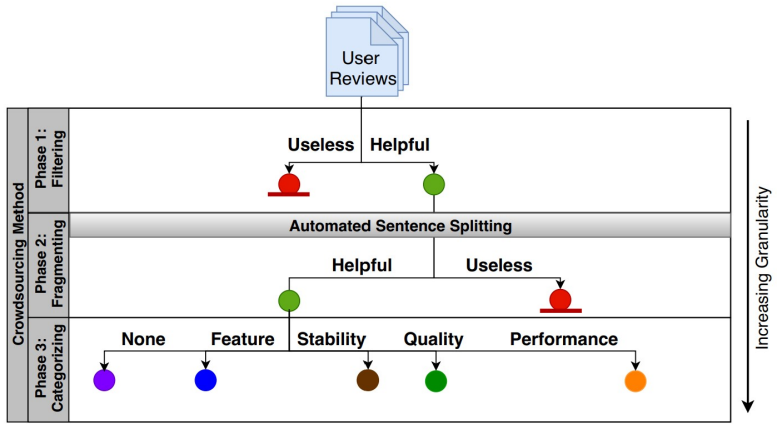}
    \caption{ Overview of the crowdsourced annotation method from Mekala et al. \cite{Mekala2021}}
    \label{fig:dataset_annotation}
\end{figure}

The dataset provided in the baseline study by Mekala et al. \cite{Mekala2021} consists of 1,000 online user reviews from the Google Play Store and the Apple App Store. This dataset was initially generated by Van Vliet et al. \cite{van2020identifying}, containing a total of 126,592 reviews, spanning across Productivity, Social Media, Messaging, and Games categories. It was further annotated through a crowdsourcing framework in three phases, P1, P2, and P3 (see Figure~\ref{fig:dataset_annotation}). However, the baseline study used only the P1 and P2 labeled datasets and published them through the Figshare repository. Below, we describe the P1 and P2 labeled datasets.
% This dataset was released via Figshare, . We refer to this dataset as baseline dataset. It included raw reviews and sentence-level labels. These reviews were crowd-annotated in three phases (see Figure~\ref{fig:dataset_annotation}), though our study only makes use of Phase 1 (P1) and Phase 2 (P2). Specifically:

\begin{itemize}
    \item \textbf{P1 (Review-level Classification):} Each review is annotated as either \textit{Helpful} (label 1) or \textit{Useless} (label 0) for software requirements engineering. These binary labels serve as ground truth for the P1 classification task.
    
    \item \textbf{P2 (Sentence-level Classification):} Reviews labeled as helpful in P1 are automatically segmented into individual sentences. Each sentence is then independently labeled as either Helpful (1) or Useless (0), forming the ground truth for the P2 task.
\end{itemize}

% \begin{figure}[ht]
%     \centering
%     \includegraphics[width=\linewidth]{figures/dataset_annotation.PNG}
%     \caption{ Overview of the crowdsourced annotation method from Mekala et al. \cite{Mekala2021}}
%     \label{fig:dataset_annotation}
% \end{figure}

\begin{figure*}[ht]
  \centering
  \includegraphics[scale=0.47]{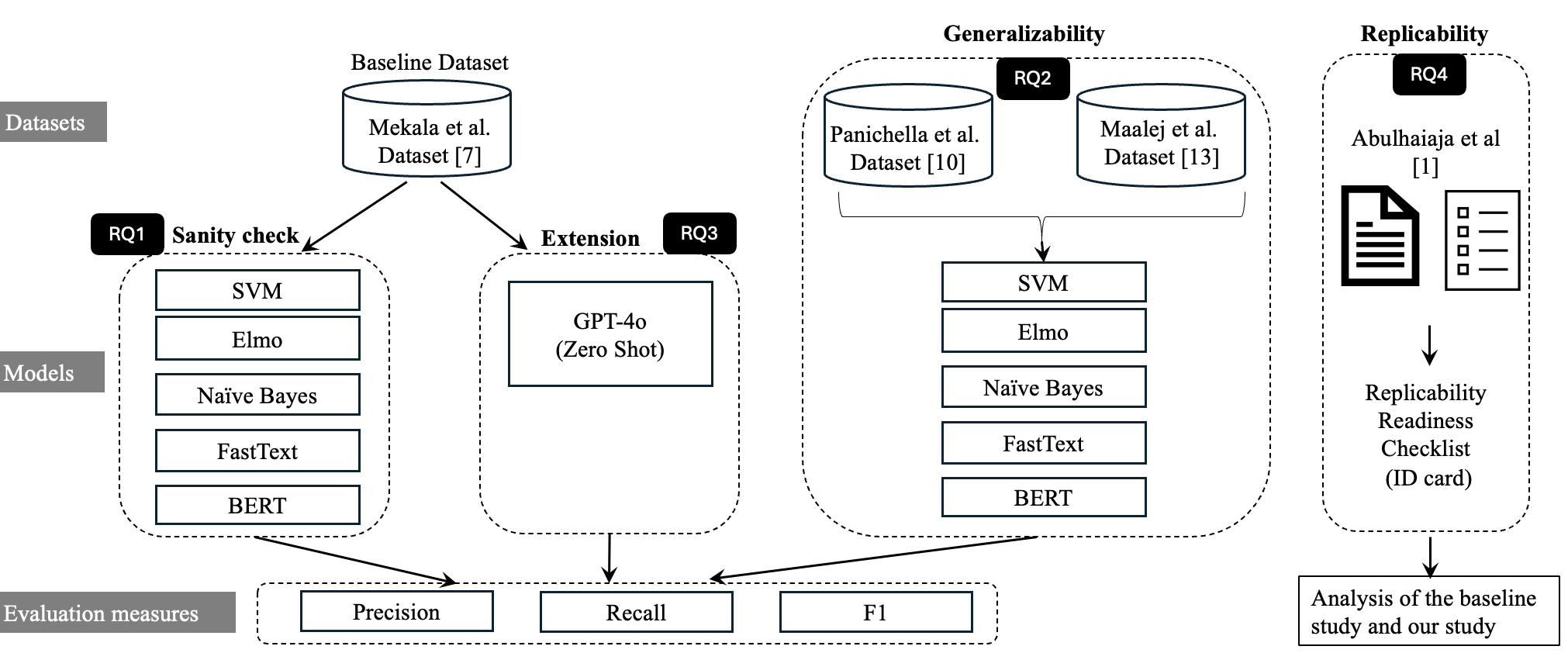}
  \caption{Our replication study design: Includes various datasets, ML models, and evaluation measures used for answering four RQs}
  \label{fig:studydesign}
\end{figure*}

% We got access to two versions of this dataset. We use these datasets to evaluate RQ1 as part of the sanity check of baseline study:

% \begin{itemize}
%     \item \textbf{baseline Dataset (Baseline 1):} This was the initial version released via Figshare, annotated through a crowdsourcing framework. We refer to this dataset as baseline dataset. It included raw reviews and sentence-level labels but contained some inconsistencies in labeling quality (as per baseline paper authors). Thus, a revised and curated dataset, referred to as the Gold Standard Dataset, was made available to us for this replication study.
    
%     \item \textbf{Gold Standard Dataset (Baseline 2):} A cleaned and manually validated version of the same dataset, provided by the authors of the baseline study. This version corrects annotation errors and inconsistencies, and includes labels for P1, P2, and P3.  However, Mekala et al.~\cite{Mekala2021} only used the P1 and P2 labels in their experiments. The P3 labels were provided for future work and categorized helpful sentences from P2 into five requirement-related types: \textit{Feature Request}, \textit{Stability Feedback}, \textit{Performance Feedback}, \textit{Quality Feedback}, and \textit{Other/None of the Above}. Our study, like theirs, focuses only on P1 and P2.
% \end{itemize}

\subsection{External Dataset }

The dataset combines two prior datasets: the Panichella \cite{panichella2015can} and Maalej Datasets \cite{maalej2016automatic}, both of which were originally developed to support supervised classification of user feedback in software engineering (see Table~\ref{tab:extended_sample}).
This dataset was accessed from the repository at Zaeem et al. \cite{Zaeem2023mohammadzaeem}. We refer to this as an external dataset, which is used to evaluate RQ2 (evaluation of the generalizability of the baseline study).

\begin{table}[ht]
\centering
\renewcommand{\arraystretch}{1.2}
\caption{Sample of the dataset from external dataset (Zaeem et al. \cite{Zaeem2023mohammadzaeem})}
\label{tab:extended_sample}
\begin{tabular}{p{5cm}p{2.5cm}}
\textbf{Review} & \textbf{Class} \\
\hline
Why limit to 50? & Information Seeking 
\\\hline

Crashes as soon as I try to load it. & Problem Discovery \\\hline

Terrible & Rating \\\hline

New interface is great & User Experience  \\\hline

I love this app. It is my go-to when I need some creative direction! & Information Giving  \\\hline
Can't read excel file correctly. After update yesterday, I can't read the email file correct as original. Please help. & Feature Request  \\

\end{tabular}
\end{table}

The \textbf{Panichella Dataset} \cite{panichella2015can} consisted of 1,390 user reviews labeled with Feature Request (FR), Problem Discovery (PD), User Experience (UE), and Rating (RT) categories. Initially, this dataset contained 32,210 reviews collected from the Google Play Store and Apple App Store, spanning across Angry Birds, Dropbox, Evernote, TripAdvisor, PicsArt, Pinterest, and WhatsApp applications. From this dataset, Panichella et al. \cite{panichella2015can} first filtered the non-informative reviews using AR-Miner \cite{chen2014ar}, and then performed the manual inspection to create the labeled dataset used in our study.

The \textbf{Maalej Dataset} \cite{maalej2016automatic} consisted of 3,691 user reviews labeled with Feature Request (FR), Problem Discovery (PD), User Experience (UE), and Rating (RT) categories. Initially, this dataset contained 1,303,182 reviews from the Google Play Store and the Apple App Store, spanning across the top four categories from the Google Play Store and all categories from the Apple App Store. From this dataset, Maalej et al. \cite{maalej2016automatic} randomly sampled 4,400 reviews, which were further labeled by 10 annotators to create a labeled dataset used in this study.

% The \textbf{Panichella Dataset} \cite{panichella2015can} consisted of 1,390 user reviews collected from apps such as Angry Birds, Dropbox, Evernote, TripAdvisor, PicsArt, Pinterest, and WhatsApp. Each review was labeled into one of four categories: Feature Request (FR), Problem Discovery (PD), Information Giving (IG), and Information Seeking (IS). The \textbf{Maalej Dataset} \cite{maalej2016automatic} included 3,691 user reviews from apps on the Google Play Store and Apple App Store, including but not limited to Dropbox, Evernote, TripAdvisor, PicsArt, Pinterest, and WhatsApp. These reviews were labeled using four classes: Feature Request (FR), Problem Discovery (PD), User Experience (UE), and Rating (RT).

Zaeem et al. \cite{Zaeem2023mohammadzaeem} merged both datasets to create a combined dataset of 6 classes. These are: Feature Request (FR), Problem Discovery (PD), Rating (RT), Information Seeking(IS), User Experience (UE), and Information Giving (IG).

Since our study focuses on binary classification for the P1 task (Helpful vs. Useless reviews), we regrouped the classes based on their utility for requirements elicitation and software development.

\textbf{Rationale for Regrouping:} Our binary classification approach is grounded in the frameworks established by Panichella et al. \cite{panichella2015can} and Maalej et al. \cite{maalej2016automatic}, who emphasize distinguishing between reviews that provide actionable feedback for software maintenance and evolution versus those that offer limited development insights. Panichella et al.\cite{panichella2015can} specifically focused their taxonomy on categories ``relevant to software maintenance and evolution,'' while Maalej et al.\cite{maalej2016automatic} noted that many reviews are ``rather non-informative, just praising the app and repeating the star ratings in words.''

% \begin{itemize}
%     \item \textbf{Helpful (1):} Reviews categorized as FR, PD, or UE. These typically contain actionable feedback, such as feature suggestions, bug reports, or user experience evaluations, that are directly relevant to requirements elicitation.
%     \item \textbf{Useless (0):} Reviews labeled as IS, IG, or RT. These include general statements, ratings, or user questions, which provide little insight into what features should be added, changed, or improved, and are therefore not useful for deriving software requirements.
% \end{itemize}

\begin{itemize}
    \item \textbf{Helpful (1):} Reviews categorized as FR, PD, or UE. These categories provide directly actionable feedback for development teams. Panichella et al.\cite{panichella2015can} define Feature Requests as ``sentences expressing ideas, suggestions or needs for improving or enhancing the app'' and Problem Discovery as ``sentences describing issues with the app or unexpected behaviours''. Maalej et al.\cite{maalej2016automatic} emphasized that bug reports (Problem Discovery) are ``critical reviews'' that development teams must prioritize, while User Experience provides valuable insights into user satisfaction and app performance.
    
    \item \textbf{Useless (0):} Reviews labeled as IS, IG, or RT. According to Panichella et al.\cite{panichella2015can}, Information Seeking represents ``attempts to obtain information or help from other users or developers'' rather than providing development insights. Information Giving involves informing users about app aspects, but may not offer actionable development feedback. Ratings align with Maalej et al.\cite{maalej2016automatic}'s observation that such reviews are typically ``non-informative, just praising the app and repeating the star ratings in words", providing limited insight for deriving software requirements.
\end{itemize}

The resulting dataset contains 1,906 reviews marked Helpful (1) and 3,162 as Useless (0).

To briefly summarize, two datasets were used in this study:
\begin{enumerate}
    \item \textbf{Baseline Dataset (P1 and P2)} — from Mekala et al., used for baseline replication.
    \item \textbf{External Dataset (Zaeem et al. \cite{Zaeem2023mohammadzaeem})} — used for testing model generalizability.
\end{enumerate}

Including this external dataset allowed us to examine whether models trained on one kind of user feedback (from app stores) can be applied to a different but similar context.

\section{Study Design} \label{design} 
\begin{figure*}[ht]
  \centering
  \includegraphics[scale=0.34]{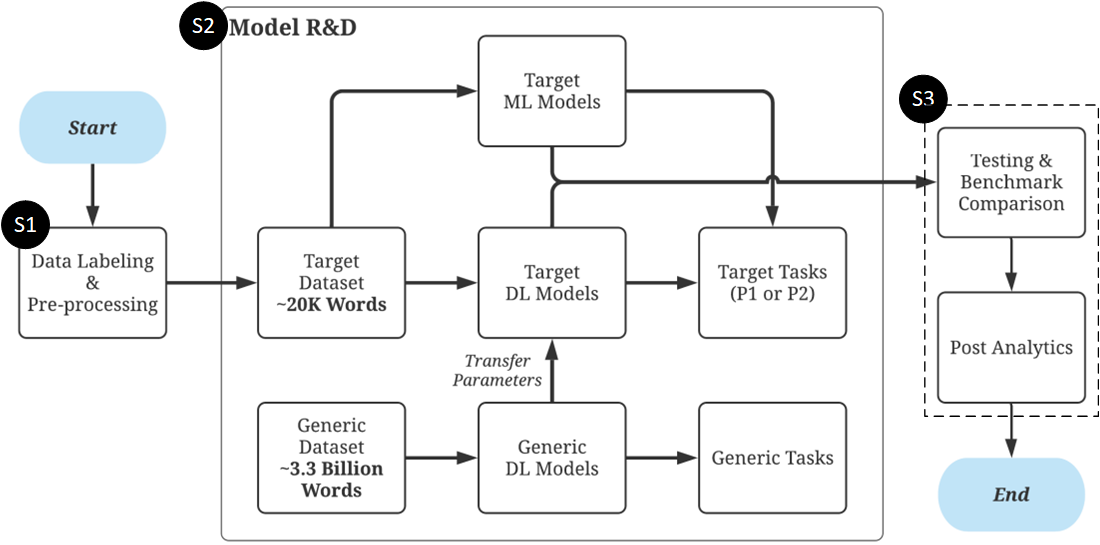}
  \caption{Baseline pipeline from Mekala et al. \cite{Mekala2021}}
  \label{fig:baseline_design}
\end{figure*}

Figure \ref{fig:studydesign} shows highlevel design of our replication study. Our study follows the baseline design (Figure \ref{fig:baseline_design}) for replicating their setup exactly as described in the baseline paper, Mekala et al.~\cite{Mekala2021}, to evaluate RQ1. We then extended this setup for additional new components to support our RQs (RQ2, RQ3 \& RQ4).

% \begin{table}[ht]
% \centering
% \caption{Summary of All Three Datasets}
% \label{tab:datasets}
% \begin{tabular}{lrrr}
% \hline
% Dataset                 & \#Reviews & \#Sentences & Helpful / Useless \\
% \hline
% Baseline baseline        & 1,000     & 1,242       & 62\% / 38\% \\
% Baseline Gold Standard  & 1,000     & 1,340       & 55\% / 45\% \\
% Extended (GitHub)       & 3,970     & —           & 48\% / 52\% \\
% \hline
% \end{tabular}
% \end{table}

\subsection{Baseline design for RQ1}
Figure~\ref{fig:baseline_design} shows the pipeline design described in the baseline study used for evaluating RQ1. It includes the following main steps:

\begin{enumerate}[label=\textbf{S\arabic*}, leftmargin=2em]
  \item \textbf{Data Labeling \& Pre-processing:} 
    User reviews for task P1 and sentences for task P2 were tokenized with special tokens (`[CLS]', `[SEP]'), mapped to numeric IDs, and padded to the maximum length using a special `[PAD]' token, with attention masks subsequently added for each of them.

  \item \textbf{Model Research \& Implementation:} 
    This step involved:
    \begin{itemize}[leftmargin=*]
      \item \textbf{Target Dataset:} The data was split with a 95:5 ratio for training and testing.
      \item \textbf{Target ML Models:} TF-IDF+SVM and Naïve Bayes classifiers were implemented as baselines.
      \item \textbf{Transfer Learning:} Three deep learning models (FastText, ELMo, and BERT) pre-trained on large public datasets were fine-tuned for tasks P1 and P2.
      \item \textbf{Training:} Models were trained for 15-25 epochs with batch size 16 and learning rates between $2 \times 10^{-5}$ and $2 \times 10^{-4}$ on a machine with 32 GB RAM, 12-core 3.50 GHz processor, and an NVidia RTX 2080 Ti GPU.
    \end{itemize}

 % \item \textbf{Training:} 
    % Models were trained for 15-25 epochs with batch size 16 and learning rates between $2 \times 10^{-5}$ and $2 \times 10^{-4}$ on a machine with 32 GB RAM, 12-core 3.50 GHz processor, and an NVidia RTX 2080 Ti GPU.

  \item \textbf{Testing \& Benchmark Comparison and Post Analytics:} 
    The trained classification models were then passed through the testing \& benchmark comparison and post analytics modules to validate the model results on unseen test data and generate detailed insights on the model performance metrics.

\end{enumerate}
% \begin{enumerate}[label=\textbf{S\arabic*}, leftmargin=2em]
%   \item \textbf{Data preprocessing:}  
%         User reviews are cleaned, tokenized, and formatted with special tokens like \texttt{[CLS]} and \texttt{[SEP]}. Padding and attention masks are added for deep learning models.
  
%   \item \textbf{Train-test split:}  
%         The data for both tasks (P1 and P2) is randomly split into 95\% training and 5\% testing. This helps make the most of the small dataset.
  
%   \item \textbf{Model training:}  
%         Five types of models are used: two traditional (TF-IDF + SVM, and Naïve Bayes), and three deep learning models (FastText, ELMo, and BERT).

%   \item \textbf{Transfer learning:}  
%         FastText, ELMo, and BERT are pre-trained on large external datasets and then fine-tuned on the requirements engineering (RE) dataset.

%   \item \textbf{Training settings:}  
%         Models are trained for 15–25 epochs with a batch size of 16. Learning rates range between $2 \times 10^{-5}$ and $2 \times 10^{-4}$. All classifiers were trained on a machine with 32 GB RAM, a 12-core 3.50 GHz processor, and an NVidia GeForce RTX 2080 Ti graphics card with 12 GB VRAM.

%   \item \textbf{Evaluation:}  
%         The test set is used to calculate precision, recall, F1 score, and accuracy.

%   \item \textbf{Post-analysis:}  
%         The baseline study also analyzed how the models performed on small amounts of data and where they made mistakes. They found that BERT improved steadily with more training data.
% \end{enumerate}

\subsection{Study design for RQ2 \& RQ3}
\textbf{Generalizability (RQ2):}
        To evaluate if the P1 models generalize well, we evaluate them, without fine-tuning, on an external dataset (from an open-source GitHub repository curated by Zaeem et al. \cite{Zaeem2023mohammadzaeem}).  We limit this extension to P1 for simplicity, since suitable public datasets are readily available at the review level, while sentence-level (P2) datasets with binary labels are difficult to find. 

        \textit{Preprocessing note for RQ2:}
        The BERT model imposes a 512-token input limit, and some entries in the external dataset exceeded this threshold, causing runtime errors. To avoid biased comparisons from truncation, we excluded these entries. All models, including those without such limits (e.g., ELMo), were evaluated on this reduced dataset (n = 5,068) to ensure consistent and fair evaluation conditions. This was only applied to the external dataset.

 \textbf{Extension (RQ3):}  
        With rapidly evolving technology,  it was pertinent to evaluate state-of-the-art models such as Generative Pre-trained Transformer models, such as GPT-4o, in a zero-shot setting. We pass raw input using the following prompt:  

        \begin{center}
        \begin{tcolorbox}[colback=gray!5!white, colframe=gray!75!black, arc=2mm, boxrule=0.5pt, fontupper=\ttfamily\small]
        Given the review, respond by saying the review is helpful or useless. If helpful, then return just '1', else just return '0'. Respond only in 1 or 0, no sentences, no words.
        \end{tcolorbox}
        \end{center}

        We started with a simpler version of the prompt: ``Given the review, respond by saying the review is helpful or useless. If helpful, return 1, else 0.'' However, the model often gave longer answers or extra text. We refined it to the final version shown above to force a clean 1/0 response. This process followed basic prompt design ideas for zero-shot settings, like those shared by Chen et al. \cite{chen2025unleashing}.  
        
        We configured the model with \texttt{temperature=0.0} and \texttt{max\_tokens=1} to ensure deterministic behavior and a strict single-token response. Each test split—P1 and P2—was evaluated five times to reduce variance, yielding a total of 22,910 predictions. GPT-4o was chosen due to its state-of-the-art language comprehension, its consistent handling of binary prompts, and its ability to perform competitively in zero-shot settings without additional fine-tuning.

% We didn’t change any training parameters, data splits, or pre-processing steps from the baseline pipeline. The same evaluation metrics (accuracy, precision, recall, F1) are used to easily compare results. 
We did not modify the training parameters, data splits, evaluation metrics (precision, recall, F1), or pre-processing steps from the baseline pipeline in order to ensure a fair and direct comparison with the baseline study.

% We used this dataset in two ways:
% \begin{itemize}
%     \item \textbf{Generalization Test:} We applied models trained on the P1 baseline dataset to this new app review dataset to evaluate how well they generalize to unseen domains.
%     \item \textbf{Zero-Shot Test:} We prompted ChatGPT-4o directly on this dataset (without fine-tuning) to assess its zero-shot classification ability.
% \end{itemize}

\subsection{Replicability evaluation - RQ4}
The ID-card is an artifact proposed by Abualhaija et al. \cite{abualhaija2024replication} to foster the replication of NLP4RE studies. It is a template composed of 47 questions concerning replication-relevant information, divided into seven topics. These topics characterize: the RE task addressed in the study; the NLP task(s) used to support the RE task; information about raw data, labeled datasets, and annotation process; implementation details; and information related to the evaluation of the proposed solution. As suggested by Abualhaija et al., we created the ID-card for the baseline study based on our understanding to support our replication study. Furthermore, we created a replication study ID-card for our study (provided in the supplementary material\footnote{\url{https://doi.org/10.5281/zenodo.15612003}}), addressing the shortcomings of the baseline study.

\subsection{Evaluation metrics}
In line with the baseline study, we employed the measures of Precision (P), Recall (R), and F1-score to compare our results with the baseline study (RQ1) and evaluate the performance of baseline models on the external dataset (RQ2) and the performance of the GPT classifier (RQ3). The F1-score (\(F1 = \frac{2*P*R}{P+R}\)) corresponds to the harmonic mean of P (\(P = \frac{TP}{TP+FP}\)) and R (\(R = \frac{TP}{TP+FN}\)), where P is the number of correct predictions out of all the input sample and R is number of positive predictions observed in the actual class. Here, True Positives (TP) refers to the number of Helpful reviews/sentences classified as Helpful, True Negatives (TN) refers to the number of Useless reviews/sentences classified as Useless, False Positives (FP) refers to the number of Useless reviews/sentences classified as Helpful, and False Negatives (FN) refers to the number to Helpful reviews/sentences classified as Useless.

\section{Results} 

\begin{table*}[]
\caption{Results answering RQ1 and RQ3: RQ1: Classification Performance of ML/DL models are directly compared with original results from baseline study \cite{Mekala2021}. RQ3: Results from GPT-4o on the baseline dataset for Tasks P1 and P2 are shown in the last row}
\label{RQ1-2 results}
\centering
\renewcommand{\arraystretch}{1.2}
\begin{tabular}{p{2.8cm}|p{2cm}||lll||lll||lll||lll|}
\multicolumn{1}{l}{{\color[HTML]{000000} \textbf{}}}                                            & {\color[HTML]{000000} \textbf{}}                                      & \multicolumn{3}{c}{{\color[HTML]{000000} \textbf{P1: Useless}}}                                                                                                    & \multicolumn{3}{c}{{\color[HTML]{000000} \textbf{P1: Helpful}}}                                                                                                     & \multicolumn{3}{c}{{\color[HTML]{000000} \textbf{P2:Useless}}}                                                                                                     & \multicolumn{3}{c}{{\color[HTML]{000000} \textbf{P2:Helpful}}}                                                                                                      \\
\multicolumn{1}{l}{{\color[HTML]{000000} \textbf{}}}                                            & {\color[HTML]{000000} \textbf{Method}}                                & {\color[HTML]{000000} \textbf{P}}                    & {\color[HTML]{000000} \textbf{R}}                    & {\color[HTML]{000000} \textbf{F1}}                   & {\color[HTML]{000000} \textbf{P}}                    & {\color[HTML]{000000} \textbf{R}}                    & {\color[HTML]{000000} \textbf{F1}}                   & {\color[HTML]{000000} \textbf{P}}                    & {\color[HTML]{000000} \textbf{R}}                    & {\color[HTML]{000000} \textbf{F1}}                   & {\color[HTML]{000000} \textbf{P}}                    & {\color[HTML]{000000} \textbf{R}}                    & {\color[HTML]{000000} \textbf{F1}}                   \\ \cline{2-14}
{\color[HTML]{000000} }                                                                         & {\color[HTML]{000000} Crowdsourcing}                                  & {\color[HTML]{000000} 0.93}                          & {\color[HTML]{000000} 0.84}                          & {\color[HTML]{000000} 0.88}                          & {\color[HTML]{000000} 0.83}                          & {\color[HTML]{000000} 0.93}                          & {\color[HTML]{000000} 0.88}                          & {\color[HTML]{000000} 0.88}                          & {\color[HTML]{000000} 0.81}                          & {\color[HTML]{000000} 0.84}                          & {\color[HTML]{000000} 0.83}                          & {\color[HTML]{000000} 0.89}                          & {\color[HTML]{000000} 0.85}                          \\
{\color[HTML]{000000} }                                                                         & {\color[HTML]{000000} \textbf{SVM}}                                   & \cellcolor[HTML]{8CC97D}{\color[HTML]{000000} 0.9}   & \cellcolor[HTML]{D5DF81}{\color[HTML]{000000} 0.79}  & \cellcolor[HTML]{BADB80}{\color[HTML]{000000} 0.84}  & \cellcolor[HTML]{BADB80}{\color[HTML]{000000} 0.83}  & \cellcolor[HTML]{92CB7D}{\color[HTML]{000000} 0.92}  & \cellcolor[HTML]{ABD37F}{\color[HTML]{000000} 0.87}  & \cellcolor[HTML]{D8DF81}{\color[HTML]{000000} 0.73}  & \cellcolor[HTML]{92CB7D}{\color[HTML]{000000} 0.92}  & \cellcolor[HTML]{C2D980}{\color[HTML]{000000} 0.82}  & \cellcolor[HTML]{BADB80}{\color[HTML]{000000} 0.83}  & \cellcolor[HTML]{FFEB84}{\color[HTML]{000000} 0.56}  & \cellcolor[HTML]{E3E382}{\color[HTML]{000000} 0.67}  \\
{\color[HTML]{000000} }                                                                         & {\color[HTML]{000000} \textbf{Naive Bayes}}                           & \cellcolor[HTML]{BADB80}{\color[HTML]{000000} 0.83}  & \cellcolor[HTML]{D5DF81}{\color[HTML]{000000} 0.79}  & \cellcolor[HTML]{CADB80}{\color[HTML]{000000} 0.81}  & \cellcolor[HTML]{CADB80}{\color[HTML]{000000} 0.81}  & \cellcolor[HTML]{B3D57F}{\color[HTML]{000000} 0.85}  & \cellcolor[HTML]{BADB80}{\color[HTML]{000000} 0.83}  & \cellcolor[HTML]{D2DD81}{\color[HTML]{000000} 0.75}  & \cellcolor[HTML]{C2D980}{\color[HTML]{000000} 0.82}  & \cellcolor[HTML]{D5DF81}{\color[HTML]{000000} 0.79}  & \cellcolor[HTML]{F2E783}{\color[HTML]{000000} 0.63}  & \cellcolor[HTML]{FFDD82}{\color[HTML]{000000} 0.52}  & \cellcolor[HTML]{FFE683}{\color[HTML]{000000} 0.57}  \\
{\color[HTML]{000000} }                                                                         & {\color[HTML]{000000} \textbf{FastText}}                              & \cellcolor[HTML]{D2DD81}{\color[HTML]{000000} 0.75}  & \cellcolor[HTML]{F9E983}{\color[HTML]{000000} 0.6}   & \cellcolor[HTML]{E3E382}{\color[HTML]{000000} 0.67}  & \cellcolor[HTML]{BADB80}{\color[HTML]{000000} 0.84}  & \cellcolor[HTML]{A0CF7E}{\color[HTML]{000000} 0.91}  & \cellcolor[HTML]{ABD37F}{\color[HTML]{000000} 0.87}  & \cellcolor[HTML]{E6E382}{\color[HTML]{000000} 0.68}  & \cellcolor[HTML]{63BE7B}{\color[HTML]{000000} 1}     & \cellcolor[HTML]{CADB80}{\color[HTML]{000000} 0.81}  & \cellcolor[HTML]{63BE7B}{\color[HTML]{000000} 1}     & \cellcolor[HTML]{F8696B}{\color[HTML]{000000} 0.25}  & \cellcolor[HTML]{FC9173}{\color[HTML]{000000} 0.4}   \\
{\color[HTML]{000000} }                                                                         & {\color[HTML]{000000} \textbf{ELMo}}                                  & \cellcolor[HTML]{BADB80}{\color[HTML]{000000} 0.83}  & \cellcolor[HTML]{CDDC81}{\color[HTML]{000000} 0.8}   & \cellcolor[HTML]{C2D980}{\color[HTML]{000000} 0.82}  & \cellcolor[HTML]{CADB80}{\color[HTML]{000000} 0.81}  & \cellcolor[HTML]{BADB80}{\color[HTML]{000000} 0.84}  & \cellcolor[HTML]{C2D980}{\color[HTML]{000000} 0.82}  & \cellcolor[HTML]{D5DF81}{\color[HTML]{000000} 0.78}  & \cellcolor[HTML]{D5DF81}{\color[HTML]{000000} 0.78}  & \cellcolor[HTML]{D5DF81}{\color[HTML]{000000} 0.78}  & \cellcolor[HTML]{E6E382}{\color[HTML]{000000} 0.68}  & \cellcolor[HTML]{E6E382}{\color[HTML]{000000} 0.68}  & \cellcolor[HTML]{E6E382}{\color[HTML]{000000} 0.68}  \\
\multirow{-6}{*}{{\color[HTML]{000000} \textbf{Original results in \cite{Mekala2021}}}} & {\color[HTML]{000000} \textbf{BERT}}                                  & \cellcolor[HTML]{73C37C}{\color[HTML]{000000} 0.95}  & \cellcolor[HTML]{A3D07E}{\color[HTML]{000000} 0.88}  & \cellcolor[HTML]{92CB7D}{\color[HTML]{000000} 0.92}  & \cellcolor[HTML]{A3D07E}{\color[HTML]{000000} 0.88}  & \cellcolor[HTML]{6CC07B}{\color[HTML]{000000} 0.96}  & \cellcolor[HTML]{92CB7D}{\color[HTML]{000000} 0.92}  & \cellcolor[HTML]{81C67C}{\color[HTML]{000000} 0.93}  & \cellcolor[HTML]{81C67C}{\color[HTML]{000000} 0.93}  & \cellcolor[HTML]{81C67C}{\color[HTML]{000000} 0.93}  & \cellcolor[HTML]{A0CF7E}{\color[HTML]{000000} 0.91}  & \cellcolor[HTML]{A0CF7E}{\color[HTML]{000000} 0.91}  & \cellcolor[HTML]{A0CF7E}{\color[HTML]{000000} 0.91}  \\ \hline
\multicolumn{14}{c}{{\color[HTML]{000000} \textbf{}}} \\ \hline
{\color[HTML]{000000} }                                                                         & {\color[HTML]{000000} \textbf{SVM}}                                   & \cellcolor[HTML]{9ACE7E}{\color[HTML]{000000} 0.89}  & \cellcolor[HTML]{CDDC81}{\color[HTML]{000000} 0.8}   & \cellcolor[HTML]{BADB80}{\color[HTML]{000000} 0.84}  & \cellcolor[HTML]{D8DF81}{\color[HTML]{000000} 0.74}  & \cellcolor[HTML]{B3D57F}{\color[HTML]{000000} 0.85}  & \cellcolor[HTML]{D5DF81}{\color[HTML]{000000} 0.79}  & \cellcolor[HTML]{CDDC81}{\color[HTML]{000000} 0.8}   & \cellcolor[HTML]{F8696B}{\color[HTML]{000000} 0.19}  & \cellcolor[HTML]{FA8571}{\color[HTML]{000000} 0.31}  & \cellcolor[HTML]{E9E582}{\color[HTML]{000000} 0.66}  & \cellcolor[HTML]{6BC07B}{\color[HTML]{000000} 0.97}  & \cellcolor[HTML]{D5DF81}{\color[HTML]{000000} 0.78}  \\
{\color[HTML]{000000} }                                                                         & {\color[HTML]{000000} \textbf{Naive Bayes}}                           & \cellcolor[HTML]{BADB80}{\color[HTML]{000000} 0.83}  & \cellcolor[HTML]{D5DF81}{\color[HTML]{000000} 0.79}  & \cellcolor[HTML]{CADB80}{\color[HTML]{000000} 0.81}  & \cellcolor[HTML]{CADB80}{\color[HTML]{000000} 0.81}  & \cellcolor[HTML]{B3D57F}{\color[HTML]{000000} 0.85}  & \cellcolor[HTML]{BADB80}{\color[HTML]{000000} 0.83}  & \cellcolor[HTML]{D2DD81}{\color[HTML]{000000} 0.75}  & \cellcolor[HTML]{C2D980}{\color[HTML]{000000} 0.82}  & \cellcolor[HTML]{D5DF81}{\color[HTML]{000000} 0.79}  & \cellcolor[HTML]{F2E783}{\color[HTML]{000000} 0.63}  & \cellcolor[HTML]{FFDD82}{\color[HTML]{000000} 0.52}  & \cellcolor[HTML]{FFE683}{\color[HTML]{000000} 0.57}  \\
{\color[HTML]{000000} }                                                                         & {\color[HTML]{000000} \textbf{FastText}}                              & \cellcolor[HTML]{C2D980}{\color[HTML]{000000} 0.82}  & \cellcolor[HTML]{F9E983}{\color[HTML]{000000} 0.6}   & \cellcolor[HTML]{E6E382}{\color[HTML]{000000} 0.69}  & \cellcolor[HTML]{B3D57F}{\color[HTML]{000000} 0.85}  & \cellcolor[HTML]{77C47C}{\color[HTML]{000000} 0.94}  & \cellcolor[HTML]{9ACE7E}{\color[HTML]{000000} 0.89}  & \cellcolor[HTML]{E3E382}{\color[HTML]{000000} 0.71}  & \cellcolor[HTML]{73C37C}{\color[HTML]{000000} 0.95}  & \cellcolor[HTML]{CADB80}{\color[HTML]{000000} 0.81}  & \cellcolor[HTML]{C2D980}{\color[HTML]{000000} 0.82}  & \cellcolor[HTML]{FC9474}{\color[HTML]{000000} 0.38}  & \cellcolor[HTML]{FFDA82}{\color[HTML]{000000} 0.51}  \\
{\color[HTML]{000000} }                                                                         & {\color[HTML]{000000} \textbf{ELMo}}                                  & \cellcolor[HTML]{ABD37F}{\color[HTML]{000000} 0.87}  & \cellcolor[HTML]{CDDC81}{\color[HTML]{000000} 0.8}   & \cellcolor[HTML]{BADB80}{\color[HTML]{000000} 0.83}  & \cellcolor[HTML]{CADB80}{\color[HTML]{000000} 0.81}  & \cellcolor[HTML]{A3D07E}{\color[HTML]{000000} 0.88}  & \cellcolor[HTML]{B3D57F}{\color[HTML]{000000} 0.85}  & \cellcolor[HTML]{C2D980}{\color[HTML]{000000} 0.82}  & \cellcolor[HTML]{D8DF81}{\color[HTML]{000000} 0.73}  & \cellcolor[HTML]{D8DF81}{\color[HTML]{000000} 0.77}  & \cellcolor[HTML]{E3E382}{\color[HTML]{000000} 0.67}  & \cellcolor[HTML]{D2DD81}{\color[HTML]{000000} 0.76}  & \cellcolor[HTML]{E0E282}{\color[HTML]{000000} 0.7}   \\
{\color[HTML]{000000} }                                                                         & {\color[HTML]{000000} \textbf{BERT}}                                  & \cellcolor[HTML]{CADB80}{\color[HTML]{000000} 0.81}  & \cellcolor[HTML]{B3D57F}{\color[HTML]{000000} 0.85}  & \cellcolor[HTML]{BADB80}{\color[HTML]{000000} 0.83}  & \cellcolor[HTML]{BADB80}{\color[HTML]{000000} 0.83}  & \cellcolor[HTML]{D5DF81}{\color[HTML]{000000} 0.79}  & \cellcolor[HTML]{CADB80}{\color[HTML]{000000} 0.81}  & \cellcolor[HTML]{81C67C}{\color[HTML]{000000} 0.93}  & \cellcolor[HTML]{81C67C}{\color[HTML]{000000} 0.93}  & \cellcolor[HTML]{81C67C}{\color[HTML]{000000} 0.93}  & \cellcolor[HTML]{A0CF7E}{\color[HTML]{000000} 0.91}  & \cellcolor[HTML]{A0CF7E}{\color[HTML]{000000} 0.91}  & \cellcolor[HTML]{A0CF7E}{\color[HTML]{000000} 0.91}  \\ \hline
\multirow{-6}{*}{{\color[HTML]{000000} \textbf{Our replicated results}}}                               & \cellcolor[HTML]{B3CEFB}{\color[HTML]{000000} \textbf{GPT Zero-Shot}} & \cellcolor[HTML]{B3CEFB}{\color[HTML]{000000} 0.84} & \cellcolor[HTML]{B3CEFB}{\color[HTML]{000000} 0.676} & \cellcolor[HTML]{B3CEFB}{\color[HTML]{000000} 0.75} & \cellcolor[HTML]{B3CEFB}{\color[HTML]{000000} 0.716} & \cellcolor[HTML]{B3CEFB}{\color[HTML]{000000} 0.87} & \cellcolor[HTML]{B3CEFB}{\color[HTML]{000000} 0.78} & \cellcolor[HTML]{B3CEFB}{\color[HTML]{000000} 0.75}  & \cellcolor[HTML]{B3CEFB}{\color[HTML]{000000} 0.75} & \cellcolor[HTML]{B3CEFB}{\color[HTML]{000000} 0.75} & \cellcolor[HTML]{B3CEFB}{\color[HTML]{000000} 0.70} & \cellcolor[HTML]{B3CEFB}{\color[HTML]{000000} 0.67} & \cellcolor[HTML]{B3CEFB}{\color[HTML]{000000} 0.68}\\
\end{tabular}
\end{table*}
\label{results}
This section presents findings from our replication study. We begin by addressing RQ1, which focuses on validating the reproducibility of the baseline study using Tasks P1 and P2. The remaining research questions (RQ2, RQ3, and RQ4) are discussed in the following subsections.

\textbf{Answering RQ1 - Sanity check:} To answer RQ1, we evaluated both how well we could reproduce the original results and how practical it was to recreate the experimental environment. This included technical setup, model training, and performance assessment across tasks.

\subsubsection{Environment Setup and Technical Challenges}

The dataset and codebase from the baseline study were publicly available, which helped us get started. However, several issues made the setup less straightforward:

\begin{itemize}
    \item \textbf{Dependency Conflicts:} Many libraries used in the original code had been updated or deprecated. For example, changes in the HuggingFace transformers library and outdated TensorFlow Hub links for ELMo caused compatibility problems.To resolve this, we reverted to (earlier versions) compatible versions.
    
    \item \textbf{Missing Setup Files:} There was no \texttt{requirements.txt} or environment file, so we had to manually install and test each dependency until the code ran successfully.
    
    \item \textbf{Runtime Instability:} We trained models using Google Colab, which sometimes timed out or ran into memory issues, especially with larger models like BERT and ELMo.
\end{itemize}

Despite these issues, we successfully reproduced the training pipelines for both P1 and P2 using the same models.

\subsubsection{Quantitative Performance Analysis}
Table \ref{RQ1-2 results} shows precision, recall, and F1-scores for each model on both P1 and P2 datasets for the results from the baseline paper and our study on the baseline dataset. All metrics are listed in the order (Precision / Recall / F1) unless otherwise noted. The baseline dataset served as the main benchmark, as it was the primary dataset used in the baseline study. 

\textbf{Note on Crowdsourcing Comparison:} Unlike the baseline study, which aimed to compare their DL pipeline against crowdsourcing approaches to demonstrate the effectiveness of automated methods, our replication study focuses specifically on reproducing and validating their reported model performance metrics. Therefore, we do not include comparisons to the crowdsourcing baseline (first row in baseline results in Table \ref{RQ1-2 results}), as our objective is to assess the reproducibility of their pipeline rather than the relative merit compared to manual annotation methods.

\begin{itemize} 
    \item \textbf{BERT} exhibited mixed reproducibility. For P1, our results showed notable variations in both classes, with F1 scores differing substantially from the baseline. For P2, both studies achieved identical scores across all metrics, demonstrating perfect consistency.

    \item \textbf{ELMo} demonstrated consistent alignment with the baseline across both tasks. While P1 results were closely matched, P2 showed minor deviations, but overall trends remained comparable.

    \item \textbf{FastText} showed greater variability. P1 results aligned reasonably well with the baseline. However, P2 revealed divergence, particularly for the Helpful class, where recall increased from the baseline study's 0.25 to our replication's 0.38, indicating sensitivity to class distribution.

    \item \textbf{SVM} yielded mixed reproducibility. P1 results followed the baseline closely, but P2 revealed a critical issue with the Useless class, where recall dropped drastically from the baseline's 0.92 to our replication's 0.19, despite maintaining high precision.

    \item \textbf{Naive Bayes} showed excellent reproducibility across both tasks, perfectly replicating the original results for both P1 and P2 on the baseline dataset.
    
\end{itemize}

\subsubsection{Qualitative Insights and Observations}

From the experiments, we made several important observations:

\begin{itemize} 

    \item Naive Bayes was the most reproducible model, demonstrating perfect reproducibility across all metrics. ELMo maintained consistent performance trends across both tasks with only minor variations. In contrast, FastText and SVM exhibited notable variations in precision-recall balance, particularly in P2 tasks.

    \item Reproducibility patterns varied by model type and task complexity. Naive Bayes achieved perfect reproducibility across both tasks, while BERT showed mixed results with perfect P2 reproducibility but notable variations in both P1 classes. Other models showed greater sensitivity in P2 sentence-level classification. This was most evident in SVM's recall collapse for the Useless class in P2.

    \item The reproducibility challenges were not uniform across evaluation metrics. While F1 scores often remained comparable between baseline and replication, underlying precision and recall values sometimes varied substantially.

\end{itemize}

\begin{tcolorbox}[answerbox]
RQ1: We achieved varying degrees of reproducibility across different models and tasks. Naive Bayes demonstrated perfect reproducibility across all metrics, while BERT showed mixed results with perfect P2 reproducibility but notable variations in both P1 classes. Models like SVM exhibited greater sensitivity, particularly in P2 tasks. We encountered practical challenges in setting up the environment and running models on cloud GPUs. These issues didn't affect the final results but highlight that reproducibility in machine learning depends heavily on having a well-documented experimental environment.
\end{tcolorbox}

\textbf{Answering RQ2 - Generalizability} To evaluate generalizability, the trained models from our reproducibility study were tested on the external dataset. As shown in(Table~\ref{tab:models_comparison_RQ2}), results reflect the average performance across five runs. This approach was taken to ensure consistent and reliable results.
\begin{itemize}
    \item \textbf{ELMo demonstrated the most balanced generalization}, with F1 scores of 0.79 for Useless and 0.68 for Helpful, resulting in strong average performance across both classes. Its precision-recall pairs (Useless: 0.82/0.76, Helpful: 0.65/0.72) were relatively consistent, suggesting robust behavior despite the class imbalance.

    \item \textbf{BERT achieved the highest single-class performance} with an F1 score of 0.82 for Useless, but its performance dropped on the Helpful class (F1: 0.63), with notably lower recall (0.54). This reflects a generalization gap, potentially caused by the model overfitting to the majority class in the extended dataset.

    \item \textbf{SVM showed reasonable but unbalanced performance}, with F1 scores of 0.70 for Useless and 0.69 for Helpful. The model achieved high precision for Useless (0.90) but lower recall (0.58), indicating it missed many actual Useless instances.
    \item \textbf{FastText demonstrated issues with class imbalance}, with F1 scores of 0.69 for Useless and 0.68 for Helpful. The model had a lower recall for Useless (0.57) despite high precision (0.88), and struggled with precision for Helpful (0.55).
    \item \textbf{Naive Bayes exhibited similar precision-recall imbalances}, achieving F1 scores of 0.69 for Useless and 0.62 for Helpful, with high precision (0.80) but lower recall (0.61) for the Useless class.
\end{itemize}

\begin{tcolorbox}[answerbox]
RQ2: These findings show that model generalizability varied across datasets. While BERT achieved the highest performance on the Useless class (F1: 0.82), its performance dropped on the Helpful class (F1: 0.63). ELMo demonstrated the most consistent performance across both classes (0.79/0.68), though with lower peak performance than BERT. Overall, deep learning models (ELMo and BERT) generalize better across datasets compared to traditional machine learning approaches.
\end{tcolorbox}

% \usepackage{multirow}
\begin{table}[ht]
\centering
\renewcommand{\arraystretch}{1.2}
\caption{RQ2: Classification Results for Task P1, comparing various ML/DL-based models on the extended dataset}
\label{tab:models_comparison_RQ2}
% \resizebox{\linewidth}{!}
{%
\begin{tabular}{p{1.5cm}|ccc|ccc}
 
\textbf{Method} & \multicolumn{3}{c|}{\textbf{P1: Useless (0)}} & \multicolumn{3}{c}{\textbf{P1: Helpful (1)}}\\

& P & R & F1 & P & R & F1\\
\hline
SVM& \textbf{0.90} & 0.58 & 0.70 & 0.56 & \textbf{0.90} & \textbf{0.69} \\
Naive Bayes& 0.80 & 0.61 & 0.69 & 0.53 & 0.74 & 0.62\\
FastText& 0.88 & 0.57 & 0.69 & 0.55 & 0.88 & 0.68\\
Elmo& 0.82 & 0.76 & 0.79 & 0.65 & 0.72 & 0.68\\
BERT& 0.76 & \textbf{0.88} & \textbf{0.82} & 0.74 & 0.54 & 0.63 \\
\end{tabular}%
}
\end{table}

\textbf{Answering RQ3-  Extension} The performance of GPT-4o as a zero-shot classifier was evaluated on the baseline dataset (in Table~\ref{RQ1-2 results}, GPT results are color coded) without any fine-tuning. We repeated each evaluation five times to account for variability and ensure consistent, reliable results; the reported metrics represent the average performance across all runs.

\begin{itemize}

     \item \textbf{GPT-4o performed moderately well for task P1 without fine-tuning}: GPT-4o achieved F1 scores of 0.75 for the useless class and 0.78 for the helpful class on the baseline dataset, demonstrating reasonable performance despite the absence of task-specific training.

    \item \textbf{Sentence-level (P2) performance showed moderate results}: GPT-4o achieved F1 scores of 0.75 for useless and 0.68 for helpful sentences. While these results demonstrate reasonable performance, they still fall below the performance of fine-tuned models, particularly BERT.
\end{itemize}

The results reveal that GPT-4o's zero-shot capabilities are slightly less effective for fine-grained sentence-level classification compared to review-level classification, despite P2 showing better precision-recall tradeoff. For P1, GPT-4o achieved higher overall performance (F1-score) but showed precision-recall imbalances (useless: 0.84/0.676, helpful: 0.716/0.87), while P2 demonstrated better balance with perfect precision-recall alignment for useless (0.75/0.75) and minimal deviation for helpful (0.70/0.67), though at the cost of lower overall F1 scores.

\begin{tcolorbox}[answerbox]
RQ3: While GPT-4o did not surpass the fine-tuned BERT model, especially for P2 classification, it performed consistently well on P1 review-level classification and remained competitive when compared to traditional machine learning models such as SVM and Naive Bayes. This could be attributed to the lack of domain knowledge in the off-the-shelf closed-source GPT models. Thus, emphasizing the need for fine-tuning further.
\end{tcolorbox}

\textbf{Answering RQ4 - Replication-readiness: }
Table \ref{tab:id} shows the replication study ID-card for the baseline study.
The baseline study addressed the requirement classification task by classifying app reviews into helpful and useless categories. To answer questions related to the dataset, we had to refer to the study of Van Vliet et al. \cite{van2020identifying}, which provided detailed information about the dataset used by the baseline study. In addition, the baseline study did not include the environment configuration file (requirements.txt), which we have incorporated into our replication package. Ultimately, based on our evaluation of the replication study ID-card, we can conclude that the baseline study is replication-ready; however, the provision of the missing information would have further enhanced the replication readiness. We also created a replication study ID-card to support the replication of our study, which is provided as supplementary material (included in our replication package).

\begin{tcolorbox}[answerbox]
RQ4: The replication study ID-card, answering 41 questions from the template provided by Abualhajia et al. \cite{abualhaija2024replication}, shows that the baseline study is almost replication-ready.
\end{tcolorbox}

\begin{table*}[htb]
    \caption{ID-card for the baseline study. This ID-card is the template of 47 questions proposed by Abualhaija et al. \cite{abualhaija2024replication} to foster the replication of NLP4RE studies.}
    \renewcommand{\arraystretch}{1.2}
    \label{tab:id}
    \begin{tabular}{p{11cm}|p{6.5cm}}
        \textbf{Question} & \textbf{Answer} \\
        \hline
        What RE task is your study addressing? & Requirements classification \\
        What types of NLP task is your study tackling? & Classification (choose among classes) \\
        What is the input of your NLP task? & Sentences \\
        What type of classification is the study about? & Binary-single label \\
        What are the labels that can be assigned? & Useless/Helpful \\
        How many data items do you process? &  1,000 records \\
        In which year or interval of years was the data produced? & 2020 \\
        What is the source of the data? & User-generated content \\
        What is the level of abstraction of the data (not limited to requirements)? & User-level \\
        What is the format of the data? & User reviews \\
        How rigorous is the format of the data? & Unconstrained natural language \\
        What is the natural language of the data (if applicable)? & English \\
        Please list which domains your data belongs to: & Productivity, Social Media, Messaging, and Games \\
        How many different sources does your data come from? & Apple App Store and Google Play Store \\
        Is the dataset publicly available (also from other authors)? & Fully \\
        What license has been used? & No license \\
        Where is the dataset stored? & In a persistent platform with DOI \\
        Provide a URL to the dataset, if available, or to the original paper that proposed the dataset: & \url{https://zenodo.org/records/3626185} \\
        How many annotators have been involved? & 603 (Crowdsourcing) \\
        How are the entries annotated? & Multiple annotators per entry \\
        What is the average level of application domain experience of the annotators? & None or unknown \\
        Who are the annotators? & Independent annotators (crowd) \\
        How was the annotation scheme established among the annotators? & Written guidelines with definitions and examples \\
        Did the authors make the written guidelines public? & Yes \\
        Did the authors share other information that could support the annotators other than the elements to annotate? & Surrounding context \\
        Did the authors employ techniques to mitigate fatigue effects during the annotation sessions? & No \\
        What are the metrics used to measure intercoder reliability? & Other (precision, recall and F1) \\
        How were conflicts resolved? & Not resolved \\
        What is the measured agreement? & Not provided \\
        What is the type of proposed solution? & Supervised deep learning \\
        What algorithms are used in the tool? & BERT, ELMo, and FastText \\
        What has been released? & Source code \\
        What needs to be done for running the tool? & Compile and run \\
        What type of documentation has been provided alongside the tool? & README file \\
        What type of dependencies does the tool have? & Open source software/libraries \\
        How is the tool released? & In a persistent platform with DOI \\
        What license has been used? & Reuse for any purpose \\
        Where is the tool released? & \url{https://doi.org/10.6084/m9.figshare.14273594} \\
        What metrics are used to evaluate the approach(es)? & Precision, Recall, and F1-score \\
        What is the validation procedure? & Train–test split \\
        What baseline do you compare against? & Automated, but self-defined \\
        Please provide more details about the baseline you compare against, if any. & Compared against traditional ML models, including SVM and Naive Bayes. \\
        \hline
    \end{tabular}
    *We excluded 6 questions from the template that were not related to the scope of the baseline study.
\end{table*}

\section{Threats to Validity} \label{threats}

\textbf{Internal Validity:} Our strict adherence to the baseline study's preprocessing pipeline, including tokenization methods and attention masking protocols, helped maintain methodological consistency. However, subtle differences in implementation environments, such as GPU architectures and memory allocation, may pose a threat to the internal validity of our study and could introduce minor variations in model training dynamics that are difficult to eliminate in replication studies. Furthermore, the sensitivity of model performance to hyperparameter selection emerged as a notable consideration, suggesting that deep-learning approaches may require more meticulous tuning to achieve consistent results. Another potential threat to the internal validity is related to the categorization of an external dataset considered in RQ2. We grouped the original categories, FR, PD, and UE, to Helpful, and IS, IG, and RT to Useless, by understanding the definitions of each category from the corresponding studies and reading the sample reviews from each category. Furthermore, this grouping was verified by our supervisor, who possesses more than 15 years of experience in the RE domain. However, future research may explore different groupings or experiment with other similar datasets.

\textbf{External Validity:} The focus on OpenAI's GPT model with zero-shot prompting may pose threats to the external validity of our study, and we acknowledge that this can limit the generalizability of our results. We encourage future studies to explore other LLMs with few-shot prompting, including open-source LLMs such as Llama3.1, Falcon, and Mistral. Additionally, LLMs often have an implicit bias from the training data, which may make the results of our study biased. However, further study will be required to assess the bias and hallucination of LLMs for RE, which is beyond the scope of this work.

\section{Related Work} \label{rw}
\textbf{DL for requirements classification}:
In various studies, deep learning (DL) has been utilized in requirements classification research to evaluate online user feedback for requirements engineering (RE). Zhou et al. \cite{zhou2016combining} employed an ensemble of Multinomial Naive Bayes and Bayesian Network classifiers to predict bug reports in user reviews, yielding favorable outcomes. Guzman et al. \cite{guzman2015ensemble} examined user feedback across eight dimensions using multiple classifiers, achieving modest precision results ranging from 69\% to 75\% for their best models. Stanik et al. \cite{stanik2019classifying} assessed traditional machine learning algorithms against a CNN-based DL model with a pre-trained FastText embedding layer, reporting average precision and recall rates of 60\% and 64\%, respectively. Consequently, existing research employing DL for user feedback classification in RE has yet to fully showcase the technology's potential.

% In several works, DL has been applied in requirements classification research to analyze online user feedback for RE. Zhou et al. \cite{zhou2016combining} used a combination (i.e., an ensemble) of Multinomial Naive Bayes and Bayesian Network classifiers that predicted whether or not a user review contains a bug report, and they achieved good results. Guzman et al. \cite{guzman2015ensemble} analyzed user feedback across eight dimensions using Naive Bayes, Support Vector Machines, Logistic Regression, Neural Network classifiers, and ensembles of these. They attained mediocre results overall, with precision values between 69\% and 75\%, respectively, for Ensemble B and Logistic Regression. Stanik et al. \cite{stanik2019classifying} compared the performance of traditional ML algorithms to a CNN-based DL model with a pre-trained FastText embedding layer to gauge the effectiveness of these models in classifying user feedback to Problem Report, Inquiry, or Irrelevant. Their reported average precision and recall results for the best-performing models were average, with 60\% and 64\%, respectively, using DL techniques. Hence, works that have applied DL for user feedback classification for RE have not demonstrated the true potential of DL yet.

\textbf{LLMs for requirements classification}:
In the RE domain, LLMs are widely used for tasks such as requirements classification \cite{arora2024advancing} and requirements elicitation \cite{ronanki2023investigating, white2024chatgpt}. However, the application of LLMs for requirement classification from user feedback is still limited. Palmetshofer et al. \cite{palmetshofer2024optimizing} and Wei et al. \cite{wei2023zero} employed Mistral and ChatGPT models, respectively, to classify app reviews into Problem Report, Inquiry, or Irrelevant categories. However, only one study compared their results with pre-trained language models fine-tuned for this specific classification task, and their results showed that ChatGPT achieved comparable performance but did not outperform fine-tuned models. Furthermore, these studies considered only three categories that lack nuanced analysis compared to the five-category classification task in the baseline study. Therefore, we extend the baseline study by integrating GPT LLM for requirement classification and comparing its results with fine-tuned models from the baseline study.

\section{Conclusion and Future Work} \label{conclude}
In this replication study of Mekala et al. \cite{Mekala2021}, we not only evaluated the internal validity of the original/baseline study but also extended it further to test its generalizability using an external dataset. Also, we utilized a state-of-the-art GPT model for this empirical study and compared the results. Finally, we utilized Abulhaija et al.'s \cite{abualhaija2024replication} work to analyze replication readiness, thereby enabling a closer examination of the elements that could help others effectively replicate these studies in the future.

The outcomes of this study were threefold. First, it enabled novice and budding researchers to learn the nuances of research in a safe environment. Second, it facilitated external replication of the research in the NLP4RE domain by regenerating the results of the baseline study and generating a replication study ID-card for the baseline study. Third, it validated the generalizability of the baseline study on an external dataset and extended the baseline study by experimenting with the GPT-4o model.

Regarding the replication of the baseline study, we observed inconsistent reproducibility outcomes across various model architectures and classification tasks. Our analysis revealed that Naive Bayes exhibited exact reproducibility across all metrics for both tasks P1 and P2, whereas BERT attained perfect reproducibility for task P2 but displayed considerable discrepancies for task P1. Additionally, we faced several challenges in executing the original code on cloud GPU systems due to the absence of environment setup files and various program and system dependencies. This issue was further emphasized by the replication study ID-card associated with the baseline study, where we did not find setup documentation beyond the basic README file. Throughout this process, significant emphasis was placed on comprehending the full implementation and design of the study, necessitating that results be generated with the understanding that complete testing of the baseline code was not feasible. Our methodology offers a validated approach to the execution of this process, with the replication study ID-card for our study to further improve replication readiness by providing the previously missing information from the baseline study.

% Regarding the replication of the baseline study, we achieved non-uniform reproducibility results across different model architectures and classification tasks. Our findings showed Naive Bayes demonstrated perfect reproducibility across all metrics for both tasks P1 and P2, while BERT achieved perfect reproducibility for task P2 but demonstrated significant variations for task P1. Furthermore, we encountered various challenges in getting the original code working on our systems due to missing environment setup files and various program and system dependencies. This challenge is also highlighted by the replication study ID-card of the baseline study, where we did not find any setup documentation other than just the README file. In the process, much-needed emphasis was placed on learning the complete implementation and design of the study, and results had to be generated with the caveat that there was no way to fully test the original program. Our methodology provides a new, confirmed way of running this process. Furthermore, we generated the replication study ID-card for our study to further enhance replication readiness by providing the missing information in the baseline study.

Regarding the generalizability of the baseline study on an external/new dataset, our results confirmed the generalizability of the fine-tuned deep learning models provided by the baseline study. For extension of the baseline study using the GPT-4o model, our findings showed that the GPT-4o model did not outperform fine-tuned deep learning models (BERT and ELMo); however, it achieved comparable performance with fine-tuned traditional machine learning models.

Building upon this successful replication, several promising avenues for future investigation emerge which are as follows: (i) expanding the evaluation to include P3 (multi-label classification) would complete the validation of the entire pipeline and provide insights into more nuanced requirements categorization tasks; (ii) comprehensive cross-domain validation studies could establish the generalizability of the approach across different feedback channels (e.g., social media, support tickets) and application domains (e.g., health\&fitness, finance, sharing economy); (iii) exploring the trade-offs between model performance and resource requirements by investigating compressed or distilled versions of BERT (e.g., DistilBERT, TinyBERT) and open-source LLMs for deployment in resource-constrained environment.

\section*{Acknowledgement}
The first four authors of this paper are the undergraduate students who contributed equally to this research work. This research work conducted over two years (2023 to 2025) was supported by Schulich School of Engineering Education Innovation research funding, University of Calgary.  
% \section*{Acknowledgement}
% We sincerely thank the original authors for generously providing the gold standard dataset and for their responsiveness in addressing our queries during the replication process. We also acknowledge the contributions of the University of Calgary's computational resources, which made the large-scale experiments feasible.


\end{document}